\documentclass[runningheads,orivec]{llncs}
\PassOptionsToPackage{table,dvipsnames,svgnames}{xcolor}
\usepackage[T1]{fontenc}
\usepackage{silence}
\usepackage{graphicx}
\usepackage{algorithm}
\usepackage{algorithmicx}
\usepackage{amsmath}
\usepackage{algpseudocode}
\usepackage{CJKutf8}
\usepackage{subcaption}
\usepackage{footmisc}
\usepackage{float}
\usepackage{booktabs}
\usepackage{multicol}
\usepackage{multirow}
\usepackage{bbding}
\usepackage{utfsym}
\usepackage{amssymb}

\begin{document}
\title{SpectMamba: Integrating Frequency and State Space Models for Enhanced
Medical Image Detection}
\author{
  Yao Wang \inst{1}\textsuperscript{*}
  \and
  Dong Yang \inst{1,2}\textsuperscript{*}
  \and
  Zhi Qiao\inst{1}
  \and
  Wenjian Huang\inst{3}
  \and
  Liuzhi Yang\inst{2}
  \and
  Zhen Qian\inst{1}\textsuperscript{†}
}
\institute{United-Imaging Research Institute of Intelligent Imaging, Beijing, China, 
\and Nankai University, Tianjin, China,   
\and Southern University of Science and Technology, Shenzhen, China, 
}
\maketitle        
\renewcommand{\thefootnote}{}
\footnote{
  *These authors contributed equally. 
  †Corresponding author.\\
   \qquad \email{\{yao.wang; zhen.qian\}@cri-united-imaging.com}
}
\renewcommand{\thefootnote}{\arabic{footnote}}
\begin{abstract}

Abnormality detection in medical imaging is a critical task requiring both high efficiency and accuracy to support effective diagnosis. While convolutional neural networks (CNNs) and Transformer-based models are widely used, both face intrinsic challenges: CNNs have limited receptive fields, restricting their ability to capture broad contextual information, and Transformers encounter prohibitive computational costs when processing high-resolution medical images. Mamba, a recent innovation in natural language processing, has gained attention for its ability to process long sequences with linear complexity, offering a promising alternative. Building on this foundation, we present SpectMamba, the first Mamba-based architecture designed for medical image detection. A key component of SpectMamba is the Hybrid Spatial-Frequency Attention (HSFA) block, which separately learns high- and low-frequency features. This approach effectively mitigates the loss of high-frequency information caused by frequency bias and correlates frequency-domain features with spatial features, thereby enhancing the model's ability to capture global context. To further improve long-range dependencies, we propose the Visual State-Space Module (VSSM) and introduce a novel Hilbert Curve Scanning technique to strengthen spatial correlations and local dependencies, further optimizing the Mamba framework. Comprehensive experiments show that SpectMamba achieves state-of-the-art performance while being both effective and efficient across various medical image detection tasks.




\end{abstract}

\section{Introduction}

Image detection has become a cornerstone of medical image analysis, with significant research dedicated to enhancing its accuracy and applicability in clinical settings \cite{litjens2017survey}. Various methods have been proposed, such as YOLO \cite{diwan2023object}, Cascade R-CNN \cite{cai2018cascade}, DETR \cite{zhang2023dense}, and DINO \cite{zhang2022dino}, each aimed at improving detection performance. These methods can generally be classified into two categories: convolutional neural network (CNN)-based approaches and Transformer-based approaches. CNN-based methods are limited by their reliance on local receptive fields, which hinder their ability to capture global features effectively. In contrast, Transformer-based methods excel at modeling long-range dependencies through the self-attention mechanism. Yet, this advantage comes at a significant computational cost, with complexity scaling quadratically with spatial resolution. 

Among the recent innovations, the Mamba framework has garnered attention as a promising alternative. Mamba models \cite{zhu2024vision,liu2024vmambavisualstatespace} integrate state-space models \cite{gu2023mamba} to reduce the quadratic complexity typically associated with traditional Transformer architectures, achieving linear complexity instead. This is accomplished by using compressed hidden states to effectively capture long-range dependencies. Mamba has already demonstrated superior performance in natural image detection and segmentation \cite{dong2024fusion,zhu2024samba}, with results indicating its potential for medical image segmentation as well \cite{ruan2024vm}. However, despite its potential, the Mamba model faces challenges in 2D medical image detection tasks.


Two primary issues arise when applying Mamba to medical image detection. First, state-space models exhibit a frequency bias \cite{yu2024tuning}, a concept first described in over-parameterized multilayer perceptrons (MLPs) \cite{rahaman2019spectral}. Frequency bias occurs when low-frequency components are learned faster than high-frequency components. It hinders the ability to capture high-frequency details, which are crucial for radiologists who rely on fine anatomical features, particularly edges and textures, when detecting and diagnosing pathological regions \cite{wei2024ct,tang2024hf,brigham1988fast}. Second, the Mamba model, designed for 1D sequences, struggles with local dependencies in 2D images. Previous attempt to adapt Mamba to 2D tasks, such as bidirectional and cascade scans \cite{liu2024vmambavisualstatespace,zhu2024vision}, involve flattening 2D patches into 1D sequences. This transformation compromises the model’s ability to interpret spatial relationships, limiting its effectiveness in abnormality detection.


In this paper, we present SpectMamba, a Mamba-based architecture specifically designed to integrate frequency and state-space models for abnormality detection in medical images. SpectMamba integrates Hybrid Spatial-Frequency Attention (HSFA) blocks and Visual State-Space Module (VSSM). The HSFA block uses parallel convolutions to capture spatial information while employing a Low-High Frequency Domain Information Separator to enhance high-frequency components. By processing the feature map in two branches: one for spatial information and the other for high- and low-frequency separation. SpectMamba reduces the impact of frequency bias, enabling it to learn high-frequency information, which distinguishes it from prior models such as FreqMamba\cite{cao2024remote,zou2024freqmamba}.



To address the issue of local dependencies in 2D image processing \cite{hu2025zigma}, we introduce the Hilbert scan curve, which preserves locality and spatial neighborhood relationships. Unlike bidirectional and cascade scans, which flatten features into 1D and disrupt spatial relationships, the Hilbert curve \cite{chen2022hilbert} maintains local clustering and spatial proximity dependencies. Additionally, the integration of feature pyramids \cite{tian2020fcos} enables multiscale representations for lesion localization, ensuring accurate detection across varying target sizes.


We validate the effectiveness of SpectMamba on three benchmark datasets, demonstrating its superior performance compared to current state-of-the-art models. SpectMamba achieves a 1.07\% relative improvement in the mean Average Precision (mAP) for pneumonia detection on X-ray images \cite{gabruseva2020deep}, a 3.7\% increase for brain tumor detection on MRI scans \cite{menze2014multimodal,bakas2017advancing,bakas2018identifying}, and a 0.48\% improvement for bone fracture detection on X-ray images \cite{nagy2022pediatric}. Furthermore, SpectMamba significantly enhances computational efficiency, operating at twice the speed of VMamba while delivering higher accuracy. Notably, it also outperforms the ViT-L-based model with a 2.52\% improvement in mAP, despite using only 32\% of its parameter count.

In summary, our contributions are as follows.

\begin{itemize} 
\item We propose the SpectMamba model, which integrates HSFA blocks to extract spatial and high-frequency details, reducing frequency bias, and incorporates a VSSM for capture long-range memory with linear complexity.

\item For the challenge of local dependency of the 1D Mamba scan, we introduce a novel Hilbert curve scanning technique to further enhance the spatial perception.

\item Extensive experiments demonstrate that SpectMamba outperforms state-of-the-art methods in medical abnormality detection tasks. 

\end{itemize}

\section{Related Work}
\textbf{State-Space Models.} State-Space Models (SSMs) \cite{mehta2022long,wang2023selective} have gained significant attention for their efficiency in modeling long-range dependencies while maintaining linear scalability with respect to sequence length. Inspired by continuous state-space models in control systems, LSSL \cite{gu2021combining}, initialized with HiPPO \cite{gu2020hippo}, demonstrated the potential of SSMs to address long-range dependency issues. However, the high computational and memory demands of LSSL make it impractical for real-world applications. To overcome these limitations, S4 \cite{gu2021efficiently} introduced diagonal parameter normalization, improving efficiency, while further advancements integrated gated units to optimize performance \cite{gu2023mamba}. The Mamba model, a data-dependent SSM with a selection mechanism and hardware-optimized design, is claimed to outperform traditional Transformers in natural language tasks while maintaining linear scalability with input length.

Building upon these advancements, selective state-space blocks have been integrated into vision backbones to enhance visual representation learning. Notable models like Vmamba \cite{liu2024vmambavisualstatespace} and Vision Mamba \cite{zhu2024vision} incorporate SSM modules into image processing pipelines. Both models demonstrate the effectiveness of Mamba-based approaches in handling diverse visual tasks. Vision Mamba employs bidirectional scanning, while Vmamba utilizes cascade scanning with four-directional coverage, both improving task performance in natural image.

\noindent \textbf{Frequency Domain Analysis.} The Fourier Transform, a fundamental technique for converting signals from the time or spatial domain to the frequency domain, is widely used in computer vision for analyzing statistical properties and frequency-domain information. GFNet \cite{rao2021global} leverages frequency domain analysis to learn long-range spatial dependencies with log-linear complexity, capturing fine-grained image attributes. Similarly, Octave Convolution \cite{chen2019drop} and HiLo \cite{pan2022fast} separate high- and low-frequency components through convolution or attention mechanisms, enabling the capture of both fine-grained and global features. FreqMamba \cite{zou2024freqmamba} explores correlations across different frequency bands for image deraining. In medical imaging, HF-ResDif \cite{tang2024hf} employs high-frequency information to guide models in reconstructing image details.

These techniques underscore the importance of distinguishing between high- and low-frequency components to improve model performance by detecting multi-scale and fine-grained features, which is particularly beneficial for medical image analysis.

\noindent \textbf{Space-Filling Curves.} Space-filling curves, such as the Hilbert curve \cite{he2022voxel}, Z-order curve \cite{orenstein1986spatial}, and sweep curve, are fractal structures that traverse every point in a multi-dimensional space without repetition \cite{mokbel2003analysis}. Widely used in point cloud processing \cite{sun2021rsn}, they reduce dimensionality while preserving spatial topology and locality. Among them, the Hilbert curve is particularly notable for its strong locality-preserving properties, making it a promising tool for spatial data processing. In contrast, many Mamba-based models in computer vision rely on scanning techniques like bidirectional scanning (Bidi-Scan) \cite{zhu2024vision} and cascade scanning (Cascade-Scan) \cite{liu2024vmambavisualstatespace}. While these methods achieve reasonable performance \cite{shi2024transnext}, these methods struggle to align with the human eye's foveal vision, often losing critical local information and compromising fine-grained spatial details. The Hilbert curve, by preserving locality, maintains proximity between image features, enabling visual models to better simulate the human eye's focus on local regions and handle detailed information more effectively.

This paper explores integrating spatial-frequency domain and state-space models with the Hilbert curve to create a robust baseline for medical image detection. By combining these techniques, we aim to enhance local feature preservation and long-range dependency modeling, thereby improving the detection of medical abnormalities.

\begin{figure*}[t]
  \centering
  \scalebox{1}{ 
   \includegraphics[width = 1\linewidth]{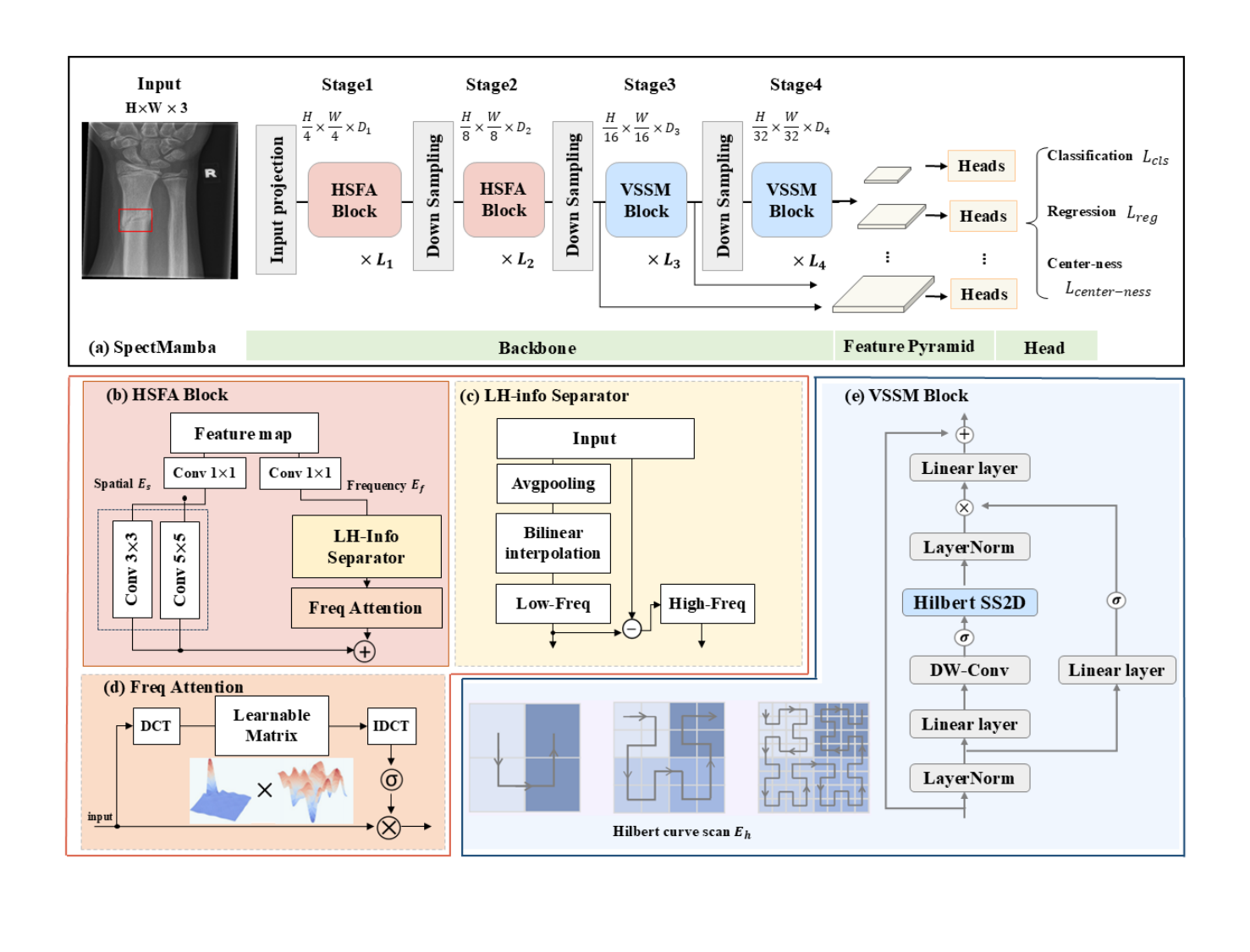}
   }
    \caption{Top: (a) The overall architecture of SpectMamba. Bottom: (b) is the structure of the Hybrid Spatial-Frequency Attention Block. (c) and (d) depict the Low-High Frequency Domain Information Separator (LH-Info Separator) and Freq attention Module, respectively. (e) presents the Visual State-Space Blocks, which incorporate the Hilbert SS2D module.}

   \label{fig:fig1}
\end{figure*}

\section{Methods}

\subsection{Preliminaries}

\subsubsection{State Space Model} In the vanilla Mamba framework, the token mixer is implemented as a selective SSM. The SSM \cite{gu2023mamba,gu2021efficiently,dao2024transformers} is a continuous-time latent state model that maps a 1D input signal $p(t) \in \mathbb{R}^{L}$ to an output signal $q(t) \in \mathbb{R}^{L}$ via a hidden state $h(t) \in \mathbb{R}^{N}$. \par
This model defines four input-dependent parameters: $(\Delta, \textbf{A}, \textbf{B}, \textbf{C})$, including a timescale parameter $\Delta$ to transform the continuous parameters A, B to discrete parameters $\overline{\textbf{A}}, \overline{\textbf{B}}$. The commonly used method for transformation is zero-order hold (ZOH), which is defined as follows:

\begin{equation}
\label{eq4}
\overline{\textbf{A}} = \exp(\Delta \textbf{A}), \quad \overline{\textbf{B}} = (\Delta \textbf{A})^{-1} \left(\exp(\Delta \textbf{A}) - \textbf{I}\right) \cdot \Delta \textbf{B}
\end{equation}
Then the SSM's sequence-to-sequence transformation is defined as:

\begin{equation}
\label{eq5}
  \begin{aligned}
    h^{\prime}(t) = \overline{\textbf{A}}h(t) + \overline{\textbf{B}}p(t), \quad q(t) = \textbf{C}h(t), 
   \end{aligned}
\end{equation}
where $\textbf{A} \in \mathbb{R}^{N \times N}$ is the evolution parameter, and $\textbf{B} \in \mathbb{R}^{N \times 1}$, $\textbf{C} \in \mathbb{R}^{1 \times N}$ are the projection parameters. Finally, the model computes the output $q$ through a global convolution. The convolutional kernel $\overline{\textbf{K}} \in \mathbb{R}^{L}$ is formulated as:

\begin{equation}
\label{eq6}
\begin{aligned}
\overline{K} &= (\textbf{C}\overline{\textbf{B}}, \textbf{C}\overline{\textbf{AB}}, \ldots, \textbf{C} \overline{\textbf{A}}^{L-1}\overline{\textbf{B}}), \\
q &= p * \overline{\textbf{K}},
\end{aligned}
\end{equation}

where $L$ is the length of the input sequence $p$, and $*$ represents the the convolution operation. This approach introduces a more efficient convolution mode that bypasses state computation, enabling robust sequence-to-sequence transformations. 

\subsection{Hybrid Spatial-Frequency Attention Block}
In Fig. \ref{fig:fig1}, we present the architecture of SpectMamba, which comprises two key modules: the Hybrid Spatial-Frequency Attention (HSFA) block and the Visual State-Space Module (VSSM). The architecture is organized into four network stages, each preceded by down-sampling layers that facilitate the creation of hierarchical feature representations. 

The HSFA block incorporates two parallel components: the Low-High Frequency Domain Information Separator (LH-Info Separator) and spatial feature extraction. The final output seamlessly integrates both spatial and frequency-domain features, enabling a more comprehensive understanding of the image.\par

Specifically, an input image $I \in \mathbb{R}^{H \times W \times C}$ is first projected and mapped to a feature map $F_{s}$. To retain the spatial information, parallel convolutions (Fig. \ref{fig:fig1}(b)) are applied to extract feature information at different scales, which is then concatenated into the feature map $Q_s$:
\begin{equation}
\label{eq7}
    \begin{aligned}
      Q_s =  \oplus_{k \in K} \{\gamma_{i}E_{s} (F_{s}, k)\},
    \end{aligned}
\end{equation}
where $E_{s}$ represents a depth-wise convolution (DW conv) with a kernel size $k$, and $\gamma_i$ is a learnable scaling factor for each DW conv. $\oplus$ denotes the concatenation operator, and $K = \{3,5\}$ is the set of kernel sizes used in this context. \par
Inspired by the high- and low-frequency separation mechanism in OctConv \cite{chen2019drop}, we employ average pooling to extract the low-frequency component, which subsamples each region through averaging rather than selectively shifting the window. This approach effectively addresses the misalignment issues commonly encountered during multi-scale information aggregation, in contrast to down-sampling via convolution with stride. The downsampling rate and step size are both set to 2, while the upsampling operation is performed using bilnear interpolation. \par
Building on this foundation, the feature vector is decomposed into low- and high-frequency components, \( F_{f}^{l} \) and \( F_{f}^{h} \),  using the FH-info module, which is designed to preserve high-frequency details.

To further enhance feature representation, we leverage the Discrete Cosine Transform (DCT), renowned for its high energy concentration and efficient conductibility, to perform domain conversion from the spatial domain to the frequency domain. Adaptive learnable matrix is then applied to the DCT-based frequency domain representation, enabling a more robust and expressive feature learning process.
\begin{equation} 
\label{eq8}
Q_{f}^{h}, Q_{f}^{l} = \text{IDCT} \big( E_{f} \big( \text{DCT}(F_{f}^{h}, F_{f}^{l}) \big) \big),
\end{equation}  
where $E_{f}$ is a learnable matrix as shown in Fig \ref{fig:fig1} (d). Finally, the HSFA block combines the spatial domain and high-low frequency domain representations as:

\begin{equation}
\label{eq91}
    \begin{aligned}
     Q_m =  \Gamma \Big( \text{Conv} \big( F_s \oplus Q_s\oplus  \big( \sigma Q_{f}^{h}\oplus \sigma Q_{f}^{l} \big) \big) \Big),
    \end{aligned}
\end{equation}
$Q_m$ is the output of the HSFA Block and $\Gamma$ represents the GELU activation function \cite{lee2023gelu}. $\oplus$ denotes the concatenation of the feature maps and $\sigma$ is the Sigmoid activation function.


\subsection{Visual State-Space Module}
After obtaining the mixed spatial and frequency domain features, the Visual State-Space Module (VSSM) performs further information fusion and feature extraction. Leveraging its long-range dependency capabilities, the VSSM captures global information effectively. However, the SSM, originally designed as a continuous-time latent state model for 1D scans, faces challenges when applied to 2D images. Previous methods, such as bidirectional and cascade scanning \cite{zhu2024vision,liu2024vmambavisualstatespace}, fail to preserve local dependencies between neighboring pixels effectively. \par

To overcome these limitations, we propose the use of the Hilbert curve as a scanning method. This method improves the capture of spatial locality by enhancing the connections between neighboring pixels \cite{he2022voxel}. As illustrated in Fig. \ref{fig:fig1}(e), Hilbert space-filling curves traverse all elements in a space without repetition or gaps, thereby preserving spatial dependencies. The mapping of the Hilbert curve of order $n$ can be represented as a mapping function from a 1D interval to a 2D space $e$: $[0, 1] \rightarrow [0, 1] \times [0, 1]$.

\begin{equation}
\label{eq9}
    E_{h}(\beta) = \lim_{n \rightarrow \infty} e_{n}(\beta), \quad \text{for } \beta \in [0, 1],
\end{equation}
Subsequently, all pixels are sorted into a 1D sequence based on their traversal position $h$:
\begin{equation}
\label{eq10}
    \begin{aligned}
        Q_{c} = E_{c}(E_{h}(Q_{m})),
    \end{aligned}
\end{equation}
where $E_{c}$ indicates the VSSM block as illustrated in Fig \ref{fig:fig1} (e), and $Q_{c}$ represents the feature vector after feature extraction by the VSSM block. Hilbert-SS2D in the VSSM block involves three steps: first, the 2D feature map is processed by traversing it along scanning paths (Hilbert-Scan); second, each sequence is independently processed by distinct S6 blocks \cite{gu2023mamba,liu2024vmambavisualstatespace}; and finally, the results are merged to generate the final 2D feature map.

\subsection{Loss Function} 
Building on feature pyramid networks (FPN) \cite{lin2017feature}, we detect objects of varying sizes across multiple levels of feature maps. Specifically, we utilize five levels of feature maps and share the detection heads across these levels. This strategy enhances parameter efficiency while simultaneously improving detection performance. \par
The loss function used during training is based on the FCOS \cite{tian2020fcos}, defined as follows:


\begin{equation}
\label{eq11}
    \begin{aligned}
    L(\{p_{x,y}\},\{t_{x,y}\},\{o_{x,y}\}) &=  \left. \frac{1}{N_{pos}}  \sum_{x,y} L_{cls}(p_{x,y}, c^{*}_{x,y}) \right. \\
    &+ \left. \frac{\lambda_1}{N_{pos}} \sum_{x,y} \mathbf{1}_{\{c^{*}_{x,y}>0\}} L_{reg}(t_{x,y}, t^{*}_{x,y}) \right. \\
    &+ \left. \frac{\lambda_2}{N_{pos}} \sum_{x,y} \mathbf{1}_{\{c^{*}_{x,y}>0\}} L_{center-ness}(o_{x,y}, o^{*}_{x,y}) \right.,   
    \end{aligned}
\end{equation}

where $L_{cls}$ is the focal loss for classification, $L_{reg}$ is the intersection over union (IoU) loss for regression, as in UnitBox \cite{yu2016unitbox} and $L_{center-ness}$ is the binary cross-entropy loss that reflects the distance from the target center. $p_{x,y}$, $t_{x,y}$, $o_{x,y}$ represent the predicted categories, positions, and center-ness at feature map location $(x,y)$, respectively, and $*$ denotes the corresponding ground truth. $ \mathbf{1}_{\{d^{*}_{x,y}>0\}}$ is equal to 1 when the specified condition is met and 0 otherwise. $N_{pos}$ denotes the number of positive samples, and $\lambda_1,\lambda_2$, set to 1 \cite{tian2022fully}.

\section{Experiments}
\subsection{Experimental Settings}
\noindent \textbf{Dataset.} 
To evaluate SpectMamba's performance in medical image detection, we conducted experiments on three widely-used public datasets from different medical domains: the RSNA Pneumonia (PenD) dataset \cite{gabruseva2020deep}, the Brain Tumor (Brats) dataset \cite{menze2014multimodal,bakas2017advancing,bakas2018identifying}, and the Bone Fracture GRAZPEDWRI-DX dataset (Graz) \cite{nagy2022pediatric}.

PenD is a public dataset for chest pneumonia detection, consisting of 6,012 clinical chest X-ray images. Brats is a MRI dataset containing 3D images of gliomas in the brain, comprising 259 cases. Following standard protocols \cite{xu2023efpn}, we sliced the 3D images along the Z axis, resulting in an average of 50 slices per case. Images in both datasets are standardized to 256×256. Graz is a dataset of 20,327 pediatric wrist trauma X-ray images, with image size standardized to 512×512. \par
We randomly split datasets into training, test and validation sets with a 70\%, 20\%, and 10\% split, respectively. To ensure fairness, the same data augmentation techniques were applied uniformly during training. All models were optimized using the adam algorithm with a weight decay of $1e^{-5}$ and trained for 200 epochs. \par
\noindent \textbf{Baseline Methods.} We evaluate SpectMamba's performance in medical image detection using three datasets. Our analysis involves two comparisons: first, we assess the impact of different backbone architectures, and second, we compare SpectMamba with state-of-the-art methods. \par
For the first comparison, models are categorized by their network structures: CNN, Transformer, and Mamba-based frameworks. CNN-based models include FCOS with ResNet \cite{tian2020fcos} and YOLOv3 with Darknet \cite{zhao2020object}. Transformer-based models include PCViT \cite{10417056}, which employs Vision Transformer (ViT) in both base and large configurations. Mamba-based models, VMamba \cite{liu2024vmambavisualstatespace} and Vision Mamba \cite{zhu2024vision}, leverage the Mamba framework for visual state-space modeling. All models are processed through FPN and trained with a consistent loss function for fair comparison. In the second comparison, we compare SpectMamba with the latest methods, including Dense Distinct Queries (DDQ) \cite{zhang2023dense} and DETR with Improved deNoising anchOrboxes (DINO) \cite{zhang2022dino}, which a complex architecture.

\noindent \textbf{Evaluation metrics.} The evaluation metrics include average precision (AP) and average recall (AR), computed across three IoU thresholds: 0.5, 0.6, and 0.7. To derive the mean average precision (mAP) and mean average recall (mAR), the AP and AR values are averaged across these thresholds.

\begin{table*}[t]

\centering
\caption{Results Across Three Datasets. The best results are highlighted in \textbf{bold}, and the second-best results are \underline{underlined}.}\label{table1}
\scalebox{0.79}{
   \begin{tabular}{>{\centering\arraybackslash}p{3.3cm} | >{\centering\arraybackslash}p{1.2cm} >{\centering\arraybackslash}p{1.2cm} >{\centering\arraybackslash}p{1.2cm} >{\centering\arraybackslash}p{1.2cm} >{\centering\arraybackslash}p{1.2cm} >{\centering\arraybackslash}p{1.2cm} >{\centering\arraybackslash}p{1.2cm} >{\centering\arraybackslash}p{1.2cm}}   
    \hline
    Method  &$AP_{50}$ &$AP_{60}$ &$AP_{70}$& $mAP$   &$AR_{50}$ &$AR_{60}$ &$AR_{70}$& $mAR$\\
        \hline\hline 
        \rowcolor[rgb]{0.8,0.78,0.78} 
        \multicolumn{9}{c}{\emph{PenD detection results}}  \\
        \hline
    \multicolumn{2}{c}{\emph{CNN-based model}} \\\cline{3-4}
    \hline        

    {ResNet\cite{tian2020fcos}}  & \underline{40.06} &23.71 &\underline{8.79} &24.19 & \underline{68.67} & \underline{47.35} & \underline{24.53} & \textbf{46.85}  \\
    {Darknet\cite{zhao2020object}}   	&39.19 &\underline{24.07} & \textbf{9.38} & \underline{24.21}  &49.30 &38.70 &\textbf{25.10} &37.70  \\
    \hline
    \multicolumn{2}{c}{\emph{transformer-based model}} \\
    \hline
    {Vit-B\cite{li2022exploring}}   &38.96 &20.62 &7.60 &22.39 &65.79 &42.84 &22.89 &43.84 \\
    {Vit-L\cite{li2022exploring}}   &39.17 &22.75 &6.93 &22.95 &66.58 &44.47 &22.04 &44.36 \\
    \hline
     \multicolumn{2}{c}{\emph{Mamba-based model}} \\
    \hline
    {Vim-base\cite{zhu2024vision}}   &34.49&19.46 &6.93 &20.29&62.66 &41.92 &20.99 &41.86 
 	\\
    {VMamba-base\cite{liu2024vmambavisualstatespace}} &36.49&21.68 &8.43 &22.20&65.08 &44.93 &24.20 &44.74 \\
    \hline
     \rowcolor[rgb]{0.92,0.92,0.92}
    {SpectMamba} & \textbf{43.11} & \textbf{24.71} & 7.95 & \textbf{25.26} &\textbf{70.11} & \textbf{47.61} & 22.76 & \underline{46.83} \\

    		\hline\hline
                    \rowcolor[rgb]{0.8,0.78,0.78} 
        \multicolumn{9}{c}{\emph{Brats detection results}}  \\
        \hline
            \multicolumn{2}{c}{\emph{CNN-based model}} \\
            \hline        
        
            {ResNet\cite{tian2020fcos}}  &85.29 &69.52&43.01 &65.94&92.93&79.15&55.91 &76.00  \\
            {Darknet\cite{zhao2020object}}   &59.70 &49.46 &27.96 &45.70 &77.23 &69.24 &51.96 &66.14   \\
            \hline
            \multicolumn{2}{c}{\emph{transformer-based model}} \\
            \hline
            {Vit-B\cite{li2022exploring}} &87.39  &72.71 & 49.71 &69.94&93.59&81.07&	61.48 &78.71 \\
            {Vit-L\cite{li2022exploring}}  &\underline{88.35} & \underline{74.56} &52.17 & \underline{71.69}	&93.62& \underline{82.80} &\underline{63.36} &79.93 	\\
            \hline
            \multicolumn{2}{c}{\emph{Mamba-based model}} \\
            \hline
            {Vim-base\cite{zhu2024vision}} &84.18 	&66.90 	&43.10 	&64.72 	&92.32 	&78.49 	&57.83 	&76.22 
         	\\
            {VMamba-base\cite{liu2024vmambavisualstatespace}} &87.96 	&74.43 	&\underline{52.52} &71.64 	& \underline{94.32} 	&82.64 	& \underline{63.36} 	& \underline{80.11}  \\
            \hline
            \rowcolor[rgb]{0.9,0.9,0.9}
            {SpectMamba}  	& \textbf{89.83} &\textbf{78.64} &\textbf{57.54} &\textbf{75.34} & \textbf{95.35} & \textbf{85.64} & \textbf{68.24} & \textbf{83.08}  
        \\
                		\hline\hline
                    \rowcolor[rgb]{0.8,0.78,0.78} 
        \multicolumn{9}{c}{\emph{Graz detection results}}\\
        \hline
             \multicolumn{2}{c}{\emph{CNN-based model}}  \\
            \hline        
        
    {ResNet\cite{tian2020fcos}}  	&87.77 &78.58 	&56.15 	&74.17 	&96.77 	&89.16 	&70.89 	&85.61 \\
    {Darknet\cite{zhao2020object}} &63.53 &54.82 	& 40.41	&52.92	& 70.48	&61.53 	&49.37&60.46  \\
    \hline
    \multicolumn{2}{c}{\emph{transformer-based model}} \\
    \hline
    {Vit-B\cite{li2022exploring}}  &86.45&77.39 &53.23&72.36&95.42 &87.53 &68.57 &83.84 \\
    {Vit-L\cite{li2022exploring}} &89.11&79.86&57.14&75.37&95.92&88.55&71.36&85.27  \\
    \hline
    \multicolumn{2}{c}{\emph{Mamba-based model}} \\
    \hline
    {Vim-base\cite{zhu2024vision}}  &86.17 	&76.66 	&54.09 	&72.31 	&96.27 &88.36 	&69.90 	&84.84
 	\\
    {VMamba-base\cite{liu2024vmambavisualstatespace}}    	& \textbf{90.55} & \underline{82.36} 	& \underline{59.33}  	& \underline{77.41} 	& \underline{97.24} 	& \textbf{91.42} 	& \underline{74.01} 	& \underline{87.56} \\
    \hline
    \rowcolor[rgb]{0.9,0.9,0.9}
    {SpectMamba} & \underline{90.27} & \textbf{82.37} & \textbf{61.03} & \textbf{77.89}& \textbf{97.54} & \underline{90.98} & \textbf{74.89} & \textbf{87.80}  \\
    		\hline\hline            
   \end{tabular}}
\end{table*}

\subsection{Comparison with Different Backbones}
\textbf{PenD Results.}  While ResNet excels at capturing local features, SpectMamba outperforms it by 3.05\% in AP@50 and 1.07\% in mAP. Its advantage stems from  the long-distance dependencies of VSSM, enabling it to better capture the globally diffuse nature of pneumonia. \textbf{Brats Results.} SpectMamba surpasses VMamba by 1.87\% in AP@50 and 3.7\% in mAP, owing to its spatial and hilbert scan module that enhance abnormality detection by capturing local details. \textbf{Graz Results.} SpectMamba outperforms VMamba by 0.48\% in mAP, leveraging its capacity to capture spatial and high-frequency dependencies. This strengthens local feature extraction, providing an advantage in detecting complex fracture patterns.\par


\textbf{Computational Efficiency.} As shown in Table \ref{table4}, SpectMamba achieves a higher mAP (77.89) compared to VMamba (77.41), while running twice as fast. Additionally, SpectMamba outperforms ViT-L (mAP 77.89 vs. 75.37) with only 32\% of the parameters.

\begin{table*}[!t]
\centering
\caption{Computational efficiency on Graz dataset. Throughput values are measured with an L40 GPU, following the protocol proposed in \cite{liu2021swin}. 
}\label{table4}
\scalebox{0.8}{
    \begin{tabular}{>{\centering\arraybackslash}p{3cm} | >{\centering\arraybackslash}p{1.5cm} >{\centering\arraybackslash}p{1.5cm} | >{\centering\arraybackslash}p{2cm} >{\centering\arraybackslash}p{2cm} >{\centering\arraybackslash}p{3cm} }
    \hline
    Method   & mAP  &mAR & Params(M). & FLOPs.(G) & Train Throughput	 \\
        \hline\hline   

    {ResNet\cite{tian2020fcos}}  &74.17 &85.61 & 32 &51.3 &914  \\
    {Darknet\cite{zhao2020object}} &52.92 &60.46 &62 & 49.9 &2642 \\
    \hline
    {Vit-B\cite{li2022exploring}}  &72.36 &83.84 & 94 & 77.3 & 691\\
    {Vit-L\cite{li2022exploring}}  &75.37 &85.27 &312 &226.1 &542 \\
    \hline
    {Vim-base\cite{zhu2024vision}}  &72.31 	&84.84& 45 &78.2 &230 
 	\\
    {VMamba-base\cite{liu2024vmambavisualstatespace}} & \underline{77.41} & \underline{87.56} &110 &38.0 & 136 \\
    \hline
    \rowcolor[rgb]{0.9,0.9,0.9}
    {SpectMamba} & \textbf{77.89} & \textbf{87.80} &100 & 77.5 & 275  
\\

    		\hline\hline
   \end{tabular}}
\end{table*}

\subsection{Comparison with SOTA Methods} 
Recent DETR-based methods, such as DDQ \cite{zhang2023dense} and DINO \cite{zhang2022dino}, have achieved SOTA performance in object detection for natural images. However, medical imaging presents unique challenges that differ significantly from natural image analysis. To rigorously assess SpectMamba's capabilities, we conducted comparative experiments under controlled conditions: no pre-trained models, consistent data augmentation strategies, 200 training epochs, and an identical learning rate.
As illustrated in Table \ref{table7}, SpectMamba consistently outperforms these SOTA models. This difference may stem from challenges unique to medical images, which differ from natural images. Medical images have high resolution and often contain small yet critical regions, such as lesions and masses. They also typically feature fewer objects per image and exhibit dataset imbalances, with a focus on a limited range of abnormalities. This results in a high class imbalance, with positive cases (e.g., unhealthy subjects) being less frequent than negative ones \cite{galdran2021balanced}, and fewer objects of interest overall \cite{dosovitskiy2020image}.



\begin{table*}[!b]
\centering
\caption{The results of SpectMamba, DDQ, and DETR across the three datasets.}\label{table7}
\scalebox{0.7}{
    \begin{tabular}{>{\centering\arraybackslash}p{3cm}| >{\centering\arraybackslash}p{1.1cm} >{\centering\arraybackslash}p{1.1cm} >{\centering\arraybackslash}p{1.1cm} >{\centering\arraybackslash}p{1.1cm} >{\centering\arraybackslash}p{1.1cm} >{\centering\arraybackslash}p{1.1cm} >{\centering\arraybackslash}p{1.1cm} >{\centering\arraybackslash}p{1.1cm} >{\centering\arraybackslash}p{1.1cm} >{\centering\arraybackslash}p{1.1cm}}
    \hline
    Method  &$AP_{50}$ &$AP_{60}$ &$AP_{70}$& mAP   &$AR_{50}$ &$AR_{60}$ &$AR_{70}$& mAR \\
        \hline\hline
    \rowcolor[rgb]{0.8,0.78,0.78} 
    \multicolumn{9}{c}{\emph{PenD detection results}}\\
    \hline
    {DDQ \cite{zhang2023dense}}  &13.83	&6.33	&2.24	&7.47  &35.12	&23.94	&14.13	&24.4 \\
    {DINO \cite{zhang2022dino}}  &14.26	&6.38	&1.99	&7.55 &30.92	&20.34	&11.05	&20.77 \\
    \hline
         \rowcolor[rgb]{0.9,0.9,0.9}
    {SpectMamba} & \textbf{43.11} & \textbf{24.71} & \textbf{7.95} & \textbf{25.26} &\textbf{70.11} & \textbf{47.61} & \textbf{22.76} & \textbf{46.83} \\
    
        \hline\hline 
    \rowcolor[rgb]{0.8,0.78,0.78} 
      \multicolumn{9}{c}{\emph{Brats detection results}}\\
    \hline        
    {DDQ \cite{zhang2023dense}}  & 48.02 & 29.87 & 12.75 &30.22 & 66.24	&51.27	&32.41	&49.97  \\
    {DINO \cite{zhang2022dino}}  &58.31	&43.23	&19.7 &40.41  &68.51	&58.14	&38.25	&54.97\\
    \hline\hline
        \rowcolor[rgb]{0.9,0.9,0.9}
    {SpectMamba}  	& \textbf{89.83} &\textbf{78.64} &\textbf{57.54} &\textbf{75.34} & \textbf{95.35} & \textbf{85.64} & \textbf{68.24} & \textbf{83.08}  \\

    \hline\hline
    \rowcolor[rgb]{0.8,0.78,0.78} 
    \multicolumn{9}{c}{\emph{Graz detection results}}\\
    \hline
    {DDQ \cite{zhang2023dense}}   &52.84&43.33&27.22&41.13&71.5&64.35&50.35	&62.06 	\\
    {DINO \cite{zhang2022dino}} &61.63	&45.02	&21	&42.55	&74.01	&62.37	&41.33	&59.24 \\
    \hline
    \rowcolor[rgb]{0.9,0.9,0.9}
    {SpectMamba} & \textbf{90.27} & \textbf{82.37} & \textbf{61.03} & \textbf{77.89}& \textbf{97.54} & \textbf{90.98} & \textbf{74.89} & \textbf{87.80}   \\

    		\hline\hline
   \end{tabular}
   }
\end{table*}


 Consistent with our findings, prior work \cite{xu2024understanding} has showed that standard practices from natural image processing—such as complex encoder architectures, multi-scale feature fusion, query initialization, and iterative bounding box refinement—often fail to enhance performance in medical imaging. In some cases, these techniques may even hinder detection accuracy.

These observations underscore the need to reconsider traditional approaches for transformer-based models and to explore more specialized, efficient frameworks tailored to the demands of medical imaging, such as SpectMamba.

\subsection{Ablation Study}
SpectMamba consists of three main components: Hilbert curve scanning, the LH-info block, and the spatial layer. The ablation results in Table \ref{table5} for these modules and different scanning modes. The baseline includes results from two reverse Hilbert curve scans. The network's performance improves when either the HSFA or spatial layer is integrated separately. Prior work\cite{liu2024vmambavisualstatespace} has shown that bidirectional scanning is more computationally efficient and yields better classification accuracy than cascade scanning. The results indicate that replacing Hilbert curve scanning with bidirectional scanning preserves the effectiveness and robustness of both the LH-info and spatial components.

    

\begin{table*}[t]
\centering
\caption{The results of the ablation experiment on the brats dataset. The best results are highlighted in \textbf{bold} and the second-best results are \underline{underlined}.}\label{table5}

\scalebox{0.8}{
    \begin{tabular}{ 
    >{\centering\arraybackslash}p{1.8cm} 
    >{\centering\arraybackslash}p{1.9cm}
    >{\centering\arraybackslash}p{1.7cm}
    >{\centering\arraybackslash}p{1.6cm}|  
    >{\centering\arraybackslash}p{1.5cm}  
    >{\centering\arraybackslash}p{1.5cm}  
    >{\centering\arraybackslash}p{1.5cm}  
    >{\centering\arraybackslash}p{1.5cm}}
    \hline
     Bidi-scan &Hilbert curve & LH-info & spatial  &$AP_{50}$  &mAP &$AR_{50}$&mAR \\
    \hline \hline   
       &$\checkmark$& & & 87.96 	&71.64 	& \underline{94.32} 	&80.11  \\\hline
      &$\checkmark$&$\checkmark$ & &87.73 	&71.38 	& \underline{94.32} 	&81.44  \\\hline
      &$\checkmark$& & $\checkmark$& \underline{89.64} 	&73.28 	&94.14 	&81.24  \\\hline
     $\checkmark$&&$\checkmark$ & $\checkmark$ &89.35 	& \underline{74.81} 	&94.23 	& \underline{82.58}  \\\hline
     \rowcolor[rgb]{0.9,0.9,0.9}
     &$\checkmark$  & $\checkmark$& $\checkmark$ & \textbf{89.83} 	& \textbf{75.33} 	& \textbf{95.35} 	& \textbf{83.08}
 \\
    		\hline\hline
   \end{tabular}
   }
\end{table*}


    

\subsection{Hilbert Curve Scanning Variants}

\begin{figure*}[htbp]
  \centering
  \begin{minipage}[b]{0.4\textwidth}
    \includegraphics[width=1\linewidth]{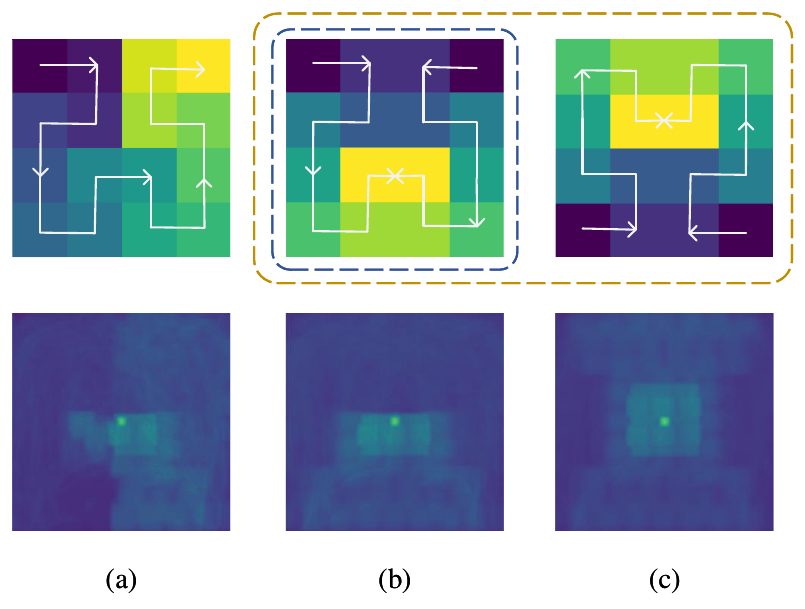}
    \caption{The figure illustrates the effective receptive fields of three Hilbert scanning methods.}
    \label{fig:fig2}
  \end{minipage}
  \hfill 
  \begin{minipage}[b]{0.57\textwidth}
    \includegraphics[width=1\linewidth]{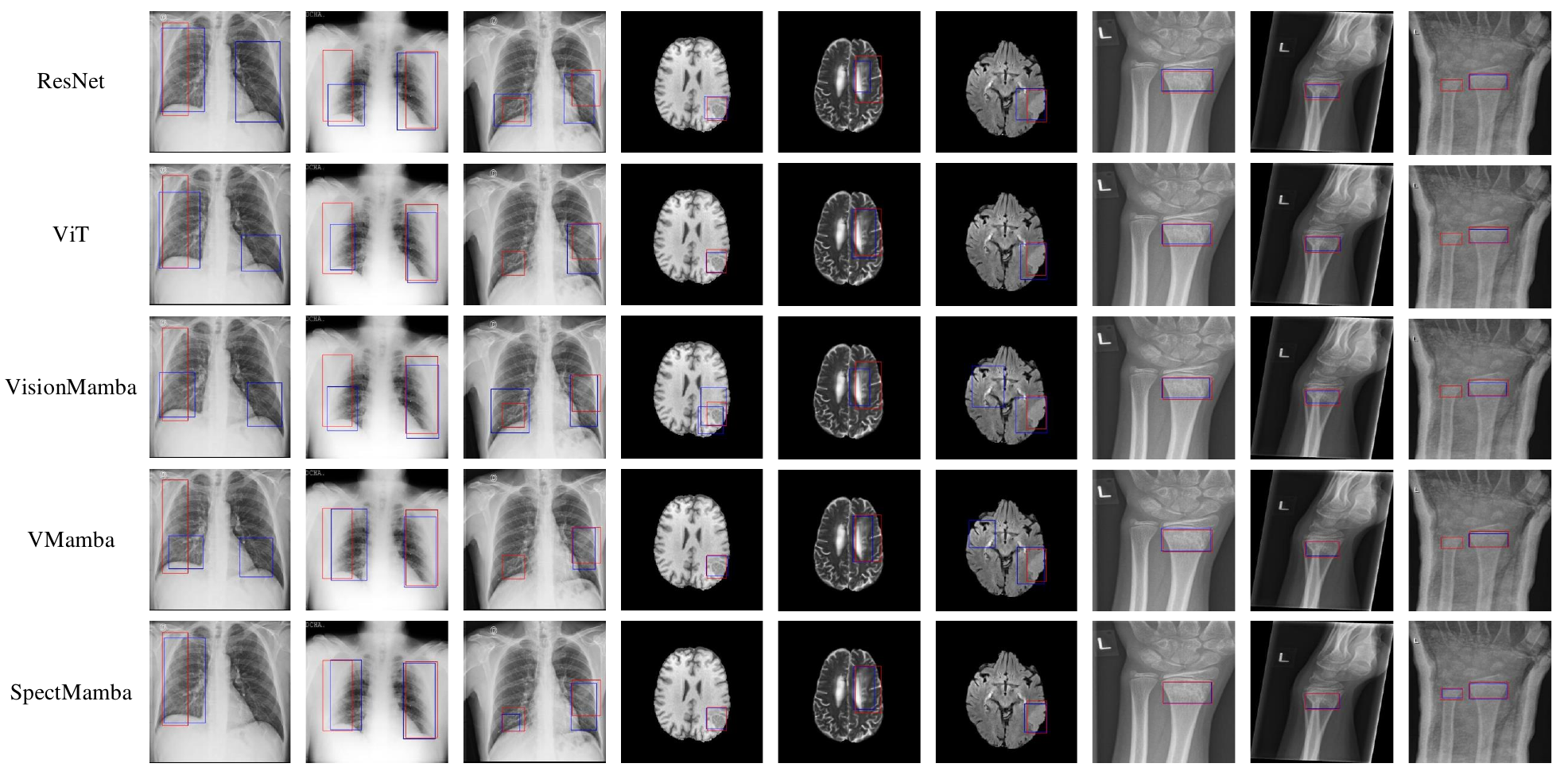}
    \caption{Visualized examples of SpectMamba and comparative methods, where red boxes indicate ground truths and blue boxes represent predictions.}
    
    \label{fig:fig4}
  \end{minipage}
  \vspace{-0.5cm}
\end{figure*}

We explored different 2D scanning paths, with Table \ref{table6} and Fig \ref{fig:fig2} presenting the results and effective receptive fields (ERFs) for each Hilbert scanning mode. The top section of Fig \ref{fig:fig2} illustrates schematic diagrams and scanning paths for the unidirectional Hilbert method, while the bottom section visualizes the ERFs \cite{luo2016understanding} for the three methods prior to training. The left image shows the unidirectional Hilbert scan curve, while the center and right images depict curves with varying start and end points.

 "Hilbert-UniDir" represents a one-way Hilbert scan, where the scan only proceeds in a forward direction without reversing, as shown in Fig. \ref{fig:fig2}(a). "Hilbert-FourDir1" involves four scanning directions: one forward and its reverse (Fig. \ref{fig:fig2}(b)), and another by flipping the matrix along the Y-axis and its reverse (Fig. \ref{fig:fig2}(c)). "Hilbert-FourDir2" also involves four scanning directions: one forward and its reverse (Fig. \ref{fig:fig2}(b)), and the other by transposing the matrix and its reverse.  "Hilbert-FourDir3" includes scanning in four directions: forward, X-axis flip, Y-axis flip, and matrix transpose.This scanning method makes the scanning path of the network symmetrical. Finally, "Hilbert-BiDir" represents scanning in both forward and reverse directions.
\begin{table*}[!t]
\centering
\caption{The results of the different Hilbert curve filling on the brats dataset. The best results are highlighted in \textbf{bold} and the second-best results are \underline{underlined}.}\label{table6}
\scalebox{0.8}{
    \begin{tabular}{ >{\centering\arraybackslash}p{3cm} |>{\centering\arraybackslash}p{1.1cm} >{\centering\arraybackslash}p{1.1cm} >{\centering\arraybackslash}p{1.1cm} >{\centering\arraybackslash}p{1.1cm} >{\centering\arraybackslash}p{1.1cm} >{\centering\arraybackslash}p{1.1cm} >{\centering\arraybackslash}p{1.1cm} >{\centering\arraybackslash}p{1.1cm}}
    \hline
     Methods &$AP_{50}$ &$AP_{60}$ &$AP_{70}$ &mAP &$AR_{50}$ &$AR_{60}$ &$AR_{70}$ &mAR\\
    \hline \hline   
     Hilbert-UniDir &88.42&75.86 &53.58 &72.62 &94.09 &83.68 &65.44 &81.07  \\\hline
     Hilbert-FourDir1&88.92& \underline{78.36} &55.54 &74.27&94.28 &84.68 &66.59&81.85  \\\hline
     Hilbert-FourDir2&89.19&77.82& \underline{56.36} & \underline{74.46} & \underline{94.66} &84.25& 
     \underline{67.55}& \underline{82.16}  \\\hline
     Hilbert-FourDir3& \underline{89.33} & 78.02&55.39& 74.25 & 94.24 & \underline{85.18} & 66.01 & 81.81  \\\hline
     \rowcolor[rgb]{0.9,0.9,0.9}
     Hilbert-BiDir& \textbf{89.83} & \textbf{78.64} & \textbf{57.54} & \textbf{75.34} & \textbf{95.35} & \textbf{85.64} & \textbf{68.24} & \textbf{83.08} 
\\\hline

    		\hline\hline
   \end{tabular}}
\end{table*}


The results show that the "Hilbert-UniDir" method already performed well, while "Hilbert-BiDir" achieved the best overall performance, outperforming all other scanning methods. Interestingly, "Hilbert-FourDir" underperformed compared to "Hilbert-BiDir," likely due to excessive overlapping paths, causing the network to overemphasize certain regions. These findings suggest that Hilbert-BiDir can replace four-directional scanning, reducing memory usage by 50\% and improving computational efficiency.



\section{Conclusion}
SpectMamba addresses the need for efficient and accurate pathology detection in medical imaging by overcoming the limitations of traditional CNNs and Transformer-based models. By leveraging Mamba's linear complexity for handling long sequences and the Hilbert curve scanning technique, the Visual State-Space Module captures global dependencies, while the Hybrid Spatial-Frequency Attention Block processes both spatial and high-frequency information. Experiments demonstrate SpectMamba's superior performance in tasks such as brain tumor detection in MRI scans and pneumonia and bone fracture detection in X-rays, highlighting its effectiveness in advancing medical diagnostics.

\bibliographystyle{splncs04}
\bibliography{paper117}
%

\end{document}